\documentclass[10pt,twocolumn,letterpaper]{article}

\usepackage{iccv}
\usepackage{times}
\usepackage{epsfig}
\usepackage{graphicx}
\usepackage{amsmath}
\usepackage{amssymb}

\graphicspath{{figures/}}

\usepackage{pifont}
\usepackage{booktabs}
\usepackage{multirow}
\usepackage{rotating}
\usepackage{algorithm}
\usepackage{algorithmic}
\usepackage{capt-of}
\usepackage{url}
\usepackage{mathtools}
\usepackage{subcaption}
\usepackage{colortbl}

\DeclareMathOperator*{\argmax}{arg\,max}

\newcommand{\Tref}[1]{Table~\ref{#1}}
\newcommand{\Eref}[1]{Eq.~(\ref{#1})}
\newcommand{\Fref}[1]{Fig.~\ref{#1}}

\newcommand{\Sref}[1]{Section~\ref{#1}}
\newcommand{\mymin}{\mathop{\rm {min}}\limits}

\newcommand{\cmark}{\ding{51}}%
\newcommand{\xmark}{\ding{55}}%
\newcommand*{\affmark}[1][*]{\textsuperscript{#1}}

\usepackage[pagebackref=true,breaklinks=true,letterpaper=true,colorlinks,bookmarks=false]{hyperref}

\iccvfinalcopy 

\def\iccvPaperID{3322} 

\ificcvfinal\fi

\begin{document}

\title{Open-Set Domain Adaptation with Visual-Language Foundation Models}

\author{Qing Yu\affmark[1]\ \ \ \ Go Irie\affmark[2]\ \ \ \ Kiyoharu Aizawa\affmark[1]\\
 \affmark[1]The University of Tokyo\ \ \ \ \affmark[2]Tokyo University of Science\\ 
{\tt\small yu@hal.t.u-tokyo.ac.jp}\ \ \ \ {\tt\small goirie@ieee.org}\ \ \ \ {\tt\small aizawa@hal.t.u-tokyo.ac.jp}
}

\maketitle
\ificcvfinal\thispagestyle{empty}\fi

\begin{abstract}
Unsupervised domain adaptation (UDA) has proven to be very effective in transferring knowledge obtained from a source domain with labeled data to a target domain with unlabeled data. Owing to the lack of labeled data in the target domain and the possible presence of unknown classes, open-set domain adaptation (ODA) has emerged as a potential solution to identify these classes during the training phase. Although existing ODA approaches aim to solve the distribution shifts between the source and target domains, most methods fine-tuned ImageNet pre-trained models on the source domain with the adaptation on the target domain. Recent visual-language foundation models (VLFM), such as Contrastive Language-Image Pre-Training (CLIP), are robust to many distribution shifts and, therefore, should substantially improve the performance of ODA. In this work, we explore generic ways to adopt CLIP, a popular VLFM, for ODA. We investigate the performance of zero-shot prediction using CLIP, and then propose an entropy optimization strategy to assist the ODA models with the outputs of CLIP. The proposed approach achieves state-of-the-art results on various benchmarks, demonstrating its effectiveness in addressing the ODA problem.
\end{abstract}

\section{Introduction}
With the increasing availability of large datasets and powerful machine learning techniques, deep learning models have achieved remarkable success in many computer vision applications, such as image recognition~\cite{he2016deep}, object detection~\cite{redmon2016you} and natural language processing~\cite{devlin2018bert}. To solve the problem of acquiring labeled large-scale data, unsupervised domain adaptation (UDA)~\cite{ghifary2016deep, taigman2016unsupervised, tzeng2017adversarial, saito2018maximum} aims to transfer knowledge learned from a labeled source domain to an unlabelled target domain. Traditional domain adaptation techniques assume that the source and target domains share the same set of classes. However, in many applications, there may exist unknown classes in the target domain that were not present in the source domain. This scenario is named open-set domain adaptation (ODA), which is a more challenging problem that addresses the transfer of knowledge across domains with different class sets, including unknown classes.

\begin{figure}[t]
    \centering
    \includegraphics[width=1\linewidth]{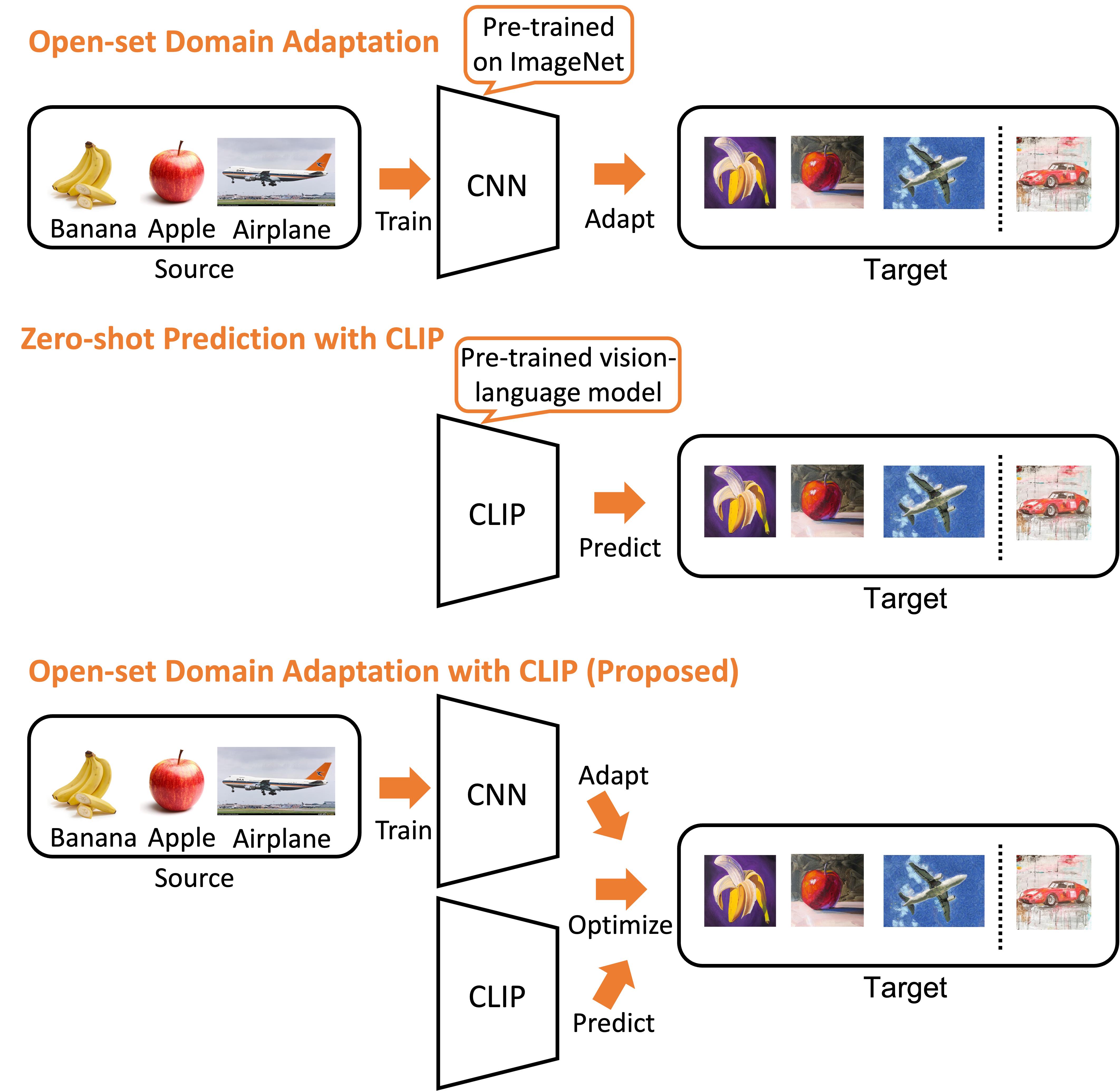}
    \caption{The conceptions of the existing ODA methods, zero-shot prediction of CLIP, and the proposed method. The proposed ODA with CLIP trains the ODA model with the guidance of CLIP.}
    \label{fig:setting}
\end{figure}

The major challenge in ODA is to identify unknown classes during the training phase. Existing ODA methods typically initialize their models with pre-trained models on ImageNet, and then fine-tune them with the source and target data, aiming to solve distribution shifts between the two domains. However, the performance of these methods heavily relies on the quality of the pre-trained models and the degree of distribution shift between the two domains.

In recent years, visual-language foundation models (VLFM), such as Contrastive Language-Image Pre-Training (CLIP)~\cite{radford2021learning}, have shown impressive performance in various computer vision and natural language processing tasks. Because these models are trained on extremely large-scale datasets containing various types of data, these models are shown to have the ability to generalize to many domains~\cite{zhang2021amortized}. Based on this kind of observation, we consider that CLIP can be used to improve the performance of ODA, including the classification of known classes and the identification of unknown classes. 

In this work, we focus on exploring the potential of CLIP for ODA. Specifically, we first investigate the robustness of CLIP for ODA on different domains and datasets. We then explore a framework to use the zero-shot predictions of CLIP to enhance the ODA performance. In our approach, we calculate the entropy of the outputs of CLIP on the target domain and the target samples having low entropy are regarded as known samples, while the target samples having high entropy are regarded as unknown samples. To achieve ODA, we train another image classification model with source samples, named ODA model. For detected known samples of the target domain, the predictions of CLIP are distilled to the ODA model, where we try to use the knowledge of CLIP to help the adaptation of target known samples. For detected unknown samples of the target domain, these samples are further separated from the known samples by maximizing the entropy of the ODA model, where the ODA model is trained to output low-confidence predictions on these unknown samples. By incorporating the outputs of CLIP with the entropy optimization strategy, we aim to provide ODA models with more informative and discriminative features, leading to better performance on ODA. 

Moreover, because the ODA model can be trained separately from the adaptation of the target domain, the coexistence of source and target samples during training is not required. This means our method can also be applied to source-free ODA (SF-ODA), where the adaptation step of target samples can be achieved only with the ODA model and no access to the source domain data is needed.

We evaluated the proposed method under various DA settings and our experimental results demonstrated that our method enhanced the ODA performance via CLIP and our technique performed far better than current ODA and SF-ODA methods. This study made the following contributions.
\begin{itemize}
    \setlength\itemsep{0pt}
    \item We investigate the performance of the zero-shot predictions obtained from CLIP in the ODA problem. 

    \item We proposed an entropy optimization strategy for the predictions of CLIP to improve ODA models in the classification of known samples and the detection of unknown samples.

    \item The proposed method can not only solve ODA, but also works in the SF-ODA setting. We evaluate our method across several benchmarks of domain adaptation and our approach outperformed other existing methods by a large margin.
    
\end{itemize}

\section{Related Work}
Currently, there are several different approaches to ODA and SF-ODA. \Tref{tbl:comp} summarizes the key methods.

\begin{table}[t]
\begin{center}
\tabcolsep = 1.2mm
\begin{tabular}{c|c|c|c}
    \toprule
     Method & \begin{tabular}[c]{@{}c@{}}Pre-trained \\ Model \end{tabular} & \begin{tabular}[c]{@{}c@{}} Source-free? \end{tabular} & \begin{tabular}[c]{@{}c@{}} Need \\ fine-tuning? \end{tabular}  \\                 \midrule
     CLIP \cite{radford2021learning} & CLIP   & \cmark & \xmark \\
     DANCE \cite{saito2020dance} & ImageNet  & \xmark & \cmark  \\ 
     OVA \cite{saito2021ovanet} & ImageNet    & \xmark & \cmark  \\
     SHOT \cite{liang2020we} & ImageNet & \cmark & \cmark  \\ 
     OneRing \cite{yang2022one} & ImageNet & \cmark & \cmark  \\ \hline
     Proposed & \begin{tabular}[c]{@{}c@{}}CLIP \\ + ImageNet  \end{tabular}  & \cmark & \cmark  \\
    \bottomrule
\end{tabular}
\end{center}
\caption{Summary of recent related methods for ODA. Our proposed method is the only method that incorporates CLIP into the ODA methods.}
\label{tbl:comp}
\end{table}

\subsection{Open-set Domain Adaptation}
Several techniques for UDA have demonstrated notable success in learning a robust classifier for labeled source data and unlabeled target data. The label sets of the source and target domains are denoted as $C_s$ and $C_t$, respectively. UDA often involves a closed-set domain adaptation task where $C_s$ equals $C_t$, and distribution alignment methods such as those proposed by~\cite{ganin2016domain, long2018conditional} have been suggested to address this task. In the presence of unknown target classes, where $C_s$ is a subset of $C_t$, ODA has been proposed as a solution to address the class mismatch problem in real-world scenarios.

One potential method for ODA is to use the importance weighting of source and target samples within a universal adaptation network, as proposed by~\cite{you2019universal}. Domain adaptive neighborhood clustering through entropy optimization (DANCE), introduced by~\cite{saito2020dance}, achieves strong performance by leveraging neighborhood clustering and entropy separation for weak domain alignment. The most advanced ODA approach is the one-versus-all network (OVANet) developed by~\cite{saito2021ovanet}, which trains one-versus-all classifiers for each class using labeled source data and adapts the open-set classifier to the target domain by minimizing the cross-entropy.

\subsection{Source-free Open-set Domain Adaptation}
It is worth noting that all prior UDA and ODA approaches require the presence of both source and target samples during training. This presents a significant challenge, as access to labeled source data may not be available after deployment for various reasons such as privacy concerns (\eg, biometric data), proprietary datasets, or simply because training on the entire source data is computationally infeasible in real-time deployment scenarios. To solve these problems, source hypothesis transfer (SHOT)~\cite{liang2020we} has been proposed for source-free UDA, which freezes the classifier module of the source model and instead focuses on learning a target-specific feature extraction module by leveraging both information maximization and self-supervised pseudo-labeling techniques. USFDA~\cite{kundu2020universal} exploits the knowledge of class-separability to detect unknown samples for SF-ODA. OneRing proposed by~\cite{yang2022one} can be adapted to the target domain easily by the weighted entropy minimization to achieve SF-ODA.

\subsection{Visual-Language Foundation Models}
With the development of Transformers for both vision ~\cite{pan2022integration, dosovitskiyimage} and language~\cite{vaswani2017attention} tasks, large-scale pre-training frameworks have become increasingly popular in recent years and have shown promising results in computer vision and natural language processing. One of the pioneer works for language pre-training is GPT~\cite{radford2018improving}, which optimizes the probability of output based on previous words in the sequence. Meanwhile, BERT~\cite{devlin2018bert} adopts the masked language modeling technique and predicts masked tokens conditioned on the unmasked ones. In computer vision, the emergence of large-scale image datasets has also led to the development of pre-training models. IGPT~\cite{chen2020generative} proposes a generative pre-training technique and shows promising results on classification tasks, while MAE~\cite{he2022masked} adopts a similar pre-training scheme as BERT and predicts the masked regions of an image with unmasked ones.

In recent years, vision-language foundation models have gained significant attention due to the availability of enormous image-text pairs collected from the internet. Various pre-training schemes have been adopted in these approaches, including contrastive learning~\cite{lu2019vilbert}, masked language modeling~\cite{su2019vl}, and masked region modeling~\cite{chen2019uniter}. CLIP~\cite{radford2021learning} is a recent representative pre-training model that aims to learn joint representations of vision and language by training on a large-scale dataset of image-text pairs. CLIP has achieved state-of-the-art performance on several visual-language benchmarks and has been shown to generalize well to different datasets. Moreover, CLIP has also been used to detect unknown samples~\cite{mingdelving}.

\begin{figure*}[t]
    \centering
    \includegraphics[width=1\linewidth]{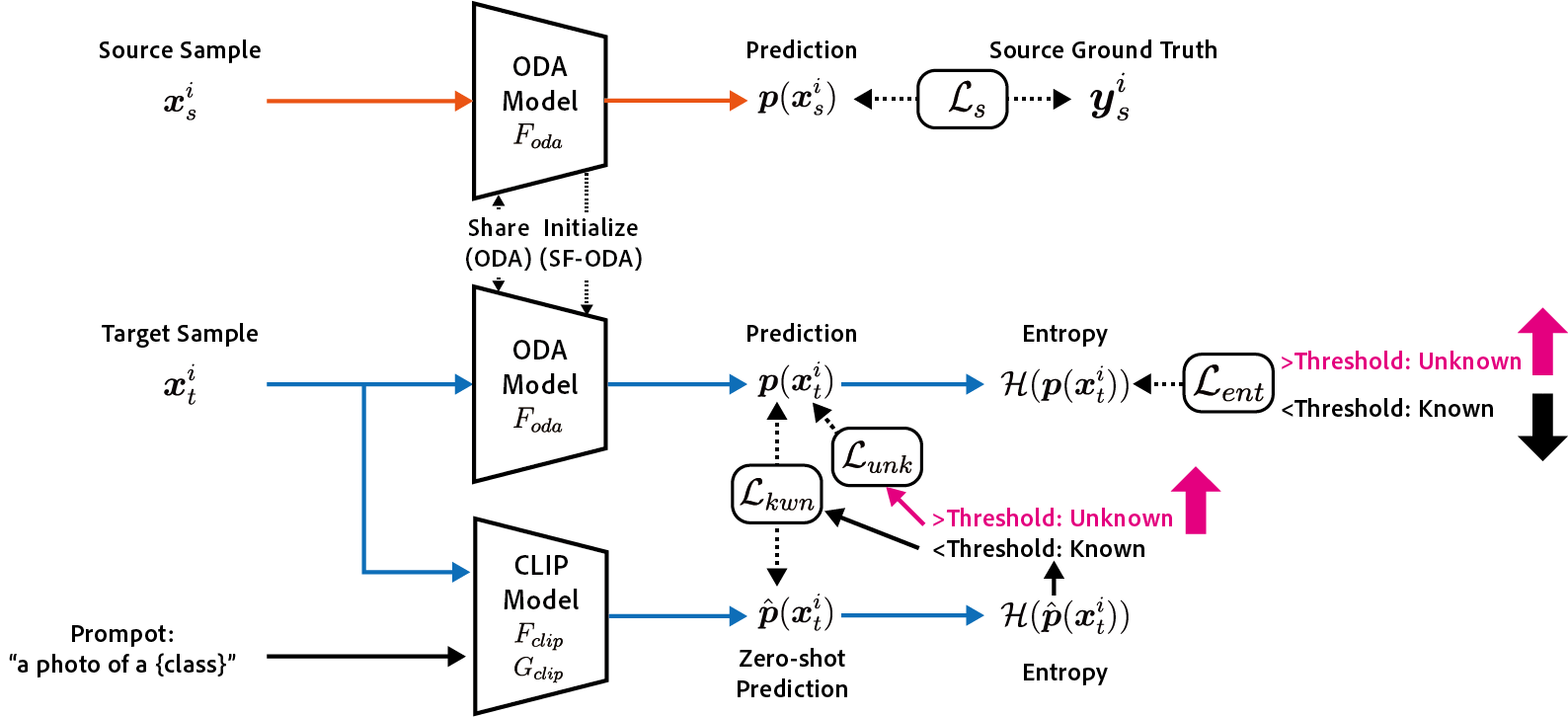}
    \caption{Overview of the proposed framework. Our network has an ODA model ($F_{oda}$) to classify the source and target samples. CLIP is also used to generate zero-shot predictions of target samples. The output of $F_{oda}$ is trained with the guidance of CLIP's predictions and entropy optimization.}
    \label{fig:method_overview}
\end{figure*}

In both ODA and SF-ODA, the existing methods usually start by initializing their models with pre-trained models on ImageNet, which is a relatively small dataset compared to the ones used in VLFM. Because the efficacy of these methods largely relies on the quality of the pre-trained models, VLFM like CLIP has a large potential to improve the performance of ODA and SF-ODA. Instead of fine-tuning VLFM with a large computational cost, we propose a lightweight way to apply the CLIP for ODA by simply using the zero-shot predictions of CLIP.

\section{Method}
In this section, we present our problem statement and proposed entropy optimization framework with CLIP for ODA and SF-ODA as shown in~\Fref{fig:method_overview}.

\subsection{Problem Statement}
\label{sec:problem}
We assume that a source image-label pair $\{\boldsymbol{x}_{s},\boldsymbol{y}_{s}\}$ is drawn from a set of labeled source images, $\{X_{s}, Y_{s}\}$, while an unlabeled target image $\boldsymbol{x}_t$ is drawn from a set of unlabeled images $X_{t}$. $C_{s}$ and $C_{t}$ denote the label sets of the source samples and target samples, respectively. In ODA, the known classes are the classes of source data, and certain unknown classes are present in the unlabeled source and target data, \ie, $C_{s} \subset C_{t}$. These unknown target classes are denoted by $\widetilde{C_t} = C_t \setminus C_{s}$. Given a target sample $\boldsymbol{x_{t}}$, the goal of ODA is to predict its label $\boldsymbol{y_{t}}$ as one of the source classes $C_{s}$ correctly or detect it as an unknown sample if it belongs to $\widetilde{C}_{t}$. In SF-ODA, $\{X_{s}, Y_{s}\}$ is not accessible when training with $X_{t}$, so the adaptation needs to be achieved only with $X_{t}$ and the model trained on  $\{X_{s}, Y_{s}\}$.

The mini-batch training process involves two sets of data, where $D_s = {(\boldsymbol{x}_s^i, \boldsymbol{y}_{s}^i)}^{N}_{i=1}$ represents a mini-batch of size $N$ that is sampled from the source samples, and $D_t = {(\boldsymbol{x}_t^i)}^{N}_{i=1}$ represents a mini-batch of size $N$ that is sampled from the target samples.

\subsection{Zero-shot Prediction using CLIP}
\label{sec:zero-shot}
CLIP is composed of an image encoder $F_{clip}$ and a language model $G_{clip}$. It utilizes the similarity between the embeddings of a text prompt $\boldsymbol{t}$ and image features to classify images, instead of using a classification head trained from scratch. The prediction is obtained by computing the cosine similarity between $F_{clip}(\boldsymbol{x}^i_{t})$ and $G_{clip}(\boldsymbol{t}_k)$ for class prompts $\boldsymbol{t}_k$:

\begin{equation}
\label{eq:zero-shot}
\hat{y}^i = \argmax_{k \in C_s} F_{clip}(\boldsymbol{x}^i_{t}) \cdot G_{clip}(\boldsymbol{t}_k),
\end{equation}
where $C_s$ is class categories in the source domain and $\cdot$ is cosine similarity.

To evaluate the power of the pre-trained CLIP model for ODA, we tested the zero-shot prediction of CLIP. We froze both the image encoder and the language model and replaced the class labels in each dataset with the text prompt $\boldsymbol{t}$ as ``A photo of a \{label\}''. In ODA, because there are unknown samples exist in the target domain, we detect these samples according to the entropy of the predictions. First, we transfer the cosine similarity to the probability $\hat{\boldsymbol{p}}$ as follows:

\begin{equation}
\hat{p}(k|{\boldsymbol {x}^i_t})= \frac{\exp (F_{clip}(\boldsymbol{x}^i_{t}) \cdot G_{clip}(\boldsymbol{t}_k) / \tau) }{\sum_{k=1}^K \exp(F_{clip}(\boldsymbol{x}^i_{t}) \cdot G_{clip}(\boldsymbol{t}_k) / \tau)},
\end{equation}
where $\hat{p}(k|{\boldsymbol {x}^i_t})$ denotes probability of the sample $\boldsymbol {x}^i_t$ belonging to class $k$ and the $\tau$ controls the distribution concentration degree and is set as $\tau=0.01$ in this paper.

We then calculate the entropy of $\hat{\boldsymbol{p}}$ as $\mathcal{H}(\hat{\boldsymbol{p}})$, and if $\mathcal{H}(\hat{\boldsymbol{p}})$ of a target sample $\boldsymbol{x}^i_{t}$ is larger than a threshold $\delta$, it will be predicted as the unknown class because the prediction has low confidence on all the known classes. For other known samples having small entropy, their classes will be predicted by~\Eref{eq:zero-shot}.

\subsection{ODA Model Preparation}
To improve ODA models with the help of CLIP, we first need to prepare a simple model to classify the source samples, which has a feature extractor and a classifier. This model is denoted as $F_{oda}$ and outputs a probability vector $F_{oda}(\boldsymbol{x}^i) \in \mathbb{R}^{|C_s|}$.
To classify the known categories correctly, we simply train the model using standard cross-entropy loss on labeled source data, expressed as:
\begin{equation}
   \mathcal{L}_{s}(D_s) = -{\frac{1}{N}}\sum_{i=1}^{N}\sum_{k=1}^{|C_s|}y_s^{ik}\log p(k|{\boldsymbol {x}^i_s}),
\label{eq:L_s}
\end{equation}
where $p(k|\boldsymbol{x}_s^i)$ represents the probability that sample $\boldsymbol{x}_s^i$ belongs to class $k$ predicted by the classifier, which is the $k$-th output of $F_{oda}(\boldsymbol{x}_s^i)$, and $y_s^{ik}$ denotes the binary label whether the sample belongs to class $k$.

\subsection{Entropy Optimization with CLIP}
\label{sec:es-clip}
As noted in~\cite{you2019universal, saito2020dance}, when compared to target known samples, the output of the classifier for target unknown samples is likely to have a higher entropy due to the absence of common features shared by known source classes. Building on this observation, we propose utilizing the entropy to distinguish between known and unknown samples. We apply this entropy strategy to both the outputs from the ODA model and the predictions from CLIP.

\subsubsection{Domain Adaptation via Entropy Separation}
To adapt the ODA model $F_{oda}$ to the target domain, we apply the entropy separation loss proposed by~\cite{saito2020dance} to the target samples as follows:
\begin{equation}
\mathcal{L}_{ent}(D_t)={\frac{1}{N}}\sum_{i=1}^{N}\mathcal{\tilde{L}}_{ent}(D_t)
\end{equation}
\begin{equation}
\mathcal{\tilde{L}}_{ent}(D_t)=
\begin{cases}
-|\mathcal{H}(\boldsymbol{p})-\delta| & \text{if } |\mathcal{H}(\boldsymbol{p})-\delta| > m\\
0 & \text{otherwise}
\end{cases},
\label{eq:L_ent}
\end{equation}
where $\mathcal{H}(\boldsymbol{p})$ is the entropy of $\boldsymbol{p}(\boldsymbol{x}_t^i)$, $\delta$ denotes the threshold and $m$ denotes the margin for separation. $\delta$ is set to $\frac{\log(|C_s|)}{2}$, because $\log(|C_s|)$ is the maximum value of $\mathcal{H}(\boldsymbol{p})$.

When the entropy is larger than the threshold and not in the margin area, \ie, $\mathcal{H}(\boldsymbol{p}) > \delta + m$ (or $\mathcal{H}(\boldsymbol{p}) - \delta > m$), this sample will be considered as an unknown sample and its entropy will be increased by minimizing~\Eref{eq:L_ent}. This step can keep the unknown samples far from the source samples. Otherwise, when the entropy is small enough, \ie, $\mathcal{H}(\boldsymbol{p}) < \delta - m$ (or $\mathcal{H}(\boldsymbol{p}) - \delta < -m$), this sample will be considered as a known sample and its entropy in~\Eref{eq:L_ent} will be decreased. This kind of entropy minimization facilitates DA of known classes in UDA tasks~\cite{carlucci2017autodial, saito2019semi}.

\subsubsection{CLIP-guided Domain Adaptation}
By the entropy separation of the ODA model, $F_{oda}$ is able to achieve ODA to a certain extent. To improve the performance of ODA, we additionally use the zero-shot predictions of CLIP to train $F_{oda}$. As mentioned in~\Sref{sec:zero-shot}, we detect the unknown samples according to the entropy of CLIP's prediction $\mathcal{H}(\hat{\boldsymbol{p}})$. We denote the detected target unknown samples as $\hat{D}_t^{unk}$ whose $\mathcal{H}(\hat{\boldsymbol{p}}) > \delta$ in $D_t$, and the remaining known samples as $\hat{D}_t^{kwn}$ whose $\mathcal{H}(\hat{\boldsymbol{p}}) <= \delta$.

For the target known samples $\hat{D}_t^{kwn}$, we directly use the zero-shot predictions of CLIP $\hat{\boldsymbol{p}}$ on these samples as the pseudo-label to train the $F_{oda}$ as follows:
\begin{equation}
   \mathcal{L}_{kwn}(\hat{D}_t^{kwn}) = -{\frac{1}{|\hat{D}_t^{kwn}|}}\sum_{i=1}^{|\hat{D}_t^{kwn}|}\sum_{k=1}^{|C_s|}\hat{p}(k|{\boldsymbol {x}^i_t})\log p(k|{\boldsymbol {x}^i_t}),
\label{eq:L_kwn}
\end{equation}
where we aim to provide ODA models with the knowledge of CLIP, leading to better classification results of known classes. 

Regarding the target unknown samples $\hat{D}_t^{unk}$, we increase the entropy of the outputs on these samples obtained from $F_{oda}$ as follows:
\begin{equation}
\mathcal{L}_{unk}(\hat{D}_t^{unk})={\frac{1}{|\hat{D}_t^{unk}|}}\sum_{i=1}^{|\hat{D}_t^{unk}|}-\mathcal{H}(\hat{\boldsymbol{p}}),
\label{eq:L_unk}
\end{equation}
where we try to incorporate the unknown classes detected by CLIP with the ones detected by $F_{oda}$ in~\Eref{eq:L_ent}.

\subsection{Overall Objective Function}
In summary, our entropy optimization framework performs the supervised training with source samples, entropy separation with target samples, and CLIP-guided domain adaptation. The overall learning objective for ODA is
\begin{equation}
\begin{aligned}
\mymin_{F_{oda}} \mathcal{L}_{total} = &\mathcal{L}_s(D_s) 
 +\mathcal{L}_{ent}(D_t) \\
 &+ \mathcal{L}_{kwn}(\hat{D}_t^{kwn})+ \mathcal{L}_{unk}(\hat{D}_t^{unk}).
\label{eq:loss}
\end{aligned}
\end{equation}

For SF-ODA, because $D_s$ is not accessible during the training with $D_t$, we assume that the ODA model $F_{oda}$ pre-trained over the source samples $D_s$ is available instead. In our experiments, we pre-train $F_{oda}$ in a standard supervised classifier learning manner, \ie, by minimizing:
\begin{equation}
\mymin_{F_{oda}} \mathcal{L}_{pretrain} = \mathcal{L}_s(D_s).
\end{equation}
We further train it with entropy separation and CLIP-guided domain adaptation over $D_t$ as follows:
\begin{equation}
\mymin_{F_{oda}} \mathcal{L}_{total} = \mathcal{L}_{ent}(D_t) + \mathcal{L}_{kwn}(\hat{D}_t^{kwn})+ \mathcal{L}_{unk}(\hat{D}_t^{unk}).
\end{equation}

\section{Experiment}
\label{sec:exp}
\subsection{Experimental Setup}
\subsubsection{Datasets}
Following existing studies~\cite{you2019universal, saito2020dance}, we used four datasets to validate our approach. (1) Office~\cite{saenko2010adapting} consists of three domains (Amazon, DSLR, Webcam), and 21 of the total 31 classes are used in ODA~\cite{saito2018open}. (2) Office-Home~\cite{venkateswara2017deep} contains four domains (Art, Clipart, Product, and Real) and 65 classes. (3) VisDA~\cite{peng2017visda} contains two domains (Synthetic and Real) and 12 classes. (4) A subset of DomainNet~\cite{peng2019moment} contains four domains (Clipart, Real, Painting, Sketch) with 126 classes. To create the scenarios for ODA, we split the classes of each dataset according to \cite{saito2020dance}, as $|C_{s}|/|\widetilde{C_t}|=10/11$ for Office, $15/50$ for OfficeHome, $6/6$ for VisDA, $60/66$ for DomainNet. 

\Tref{tab:dataset} summarizes the overall statistics of each dataset used in our experiments. For Office and Office-Home, each domain is used as the source and target domains. For VisDA, only the synthetic-to-real task was performed. For DomainNet, seven tasks from four domains (C2S, P2C, P2R, R2C, R2P, R2S, S2P) were performed as described in~\cite{saito2019semi}. 

\begin{table}[t]
\begin{center}
\begin{tabular}{ccccc}
\toprule
Dataset & Domain & \begin{tabular}[c]{@{}c@{}}\#Total \\ known \\ samples\end{tabular} & \begin{tabular}[c]{@{}c@{}}\#Total \\ unknown \\ samples\end{tabular} \\ \hline\hline
\multirow{3}{*}{Office} & Amazon (A) & 958 & 1009 \\
 & DSLR (D) & 157  &  175  \\
 & Webcam (W) & 295 &  269 \\ \hline
 \multirow{4}{*}{ \begin{tabular}[c]{@{}c@{}}Office \\ -Home\end{tabular}} & Art (A) & 743  & 1,684   \\
 & Clipart (C) & 1,116  & 3,249   \\
 & Product (P) & 1,077  & 3,362  \\
 & Real (R) & 1,203  & 3,154  \\ \hline
 \multirow{2}{*}{VisDA} & Synthetic & 79,765 & - \\
 & Real & 34,146  & 21,242 \\ \hline
  \multirow{4}{*}{ \begin{tabular}[c]{@{}c@{}}DomainNet\end{tabular}} & Clipart (C) & 8,333 & 10,370   \\
 & Painting (P) & 13,049  & 18,453   \\
 & Real (R)& 33,238  & 37,120  \\
 & Sketch (S)& 9,309  & 15,273  \\ \bottomrule
\end{tabular}
\end{center}
\caption{Overall statistics of each dataset.}
\label{tab:dataset}
\end{table}

\subsubsection{Comparison of Methods} We compared the proposed method with two baseline methods: (1) the zero-shot prediction by CLIP~\cite{radford2021learning} as described in \Sref{sec:zero-shot}, and (2) source only (SO), in which the model is trained only with labeled source data and the unknown classes are detected based on the entropy. We also compared it with two ODA methods, (1) DANCE~\cite{saito2020dance} and (2) OVA~\cite{saito2021ovanet}, and two SF-ODA methods, (1) SHOT~\cite{liang2020we} and (2) OneRing~\cite{yang2022one}. We chose to exclude the results of standard domain alignment baselines such as DANN~\cite{ganin2016domain} and CDAN~\cite{long2018conditional} from our analysis, as prior research of ODA~\cite{saito2020dance, yang2021generalized} has demonstrated that these methods can lead to a notable decline in performance when rejecting unknown samples.

\subsubsection{Evaluation Protocols} Evaluating ODA methods requires taking into account the trade-off between the accuracy of known and unknown classes. We used the H-score metric~\cite{saito2021ovanet, bucci2020effectiveness}. When the unknown classes are regarded as a unified unknown class, the H-score is the harmonic mean of the accuracy of known classes ($\mathrm{acc}_{kwn}$) and that of the unified unknown class ($\mathrm{acc}_{unk}$), as follows. 

\begin{equation}
    H_{score} = \frac{2 \mathrm{acc}_{kwn} \cdot \mathrm{acc}_{unk}}{\mathrm{acc}_{kwn} + \mathrm{acc}_{unk}}.
\end{equation}
The H-score is high only when both $\mathrm{acc}_{kwn}$ and $\mathrm{acc}_{unk}$ are high, indicating that this metric accurately measures both accuracies. We report the averages of the scores obtained from three trials with different random seeds in all experiments.

\subsubsection{Implementation Details} All experiments are implemented in PyTorch~\cite{NEURIPS2019_9015}. We used the same network architecture and hyperparameters as in~\cite{saito2020dance}. We implemented our network using ResNet-50~\cite{he2016deep} pre-trained on ImageNet~\cite{deng2009imagenet} as the ODA model $F_{oda}$. We set the threshold of the entropy as $\frac{\log(|C_s|)}{2}$ and the margin $m$ as 0.5 for our method. For CLIP, we use the original implementation and model in~\cite{radford2021learning}.

\subsection{Experimental Results}
\label{sec:result}

\begin{table}[t]
\begin{center}
\begin{tabular}{c|cccc}
\toprule
Method &  OF  & OH & VD & DN \\\hline\hline
CLIP    & 76.37 & 65.71 & 79.49 & 66.16 \\ \hline
SO      & 60.91 & 56.88 & 43.57 & 59.14 \\
DANCE   & 77.82 & 63.33 & 67.87 & 58.03 \\
OVA     & 89.57 & 70.61 & 59.80 & 62.23 \\
\rowcolor[gray]{0.90}Ours    & \textbf{92.79} & 79.43 & 80.68 & \textbf{76.23} \\ \hline
SHOT    & 78.00 & 63.08 & 47.08 & 58.64 \\
OneRing & 89.87 & 67.65 & 51.21 & 58.46 \\
\rowcolor[gray]{0.90}Ours SF & 91.87 & \textbf{80.67} & \textbf{83.81} & 76.13 \\
\bottomrule
\end{tabular}
\end{center}
\caption{H-scores (\%) of ODA and SF-ODA on each dataset (OF: Office, OH: Office-Home, VD: VisDA, DN: DomainNet). ``Ours SF'' denotes the source-free version of the proposed method. The average scores of all tasks for each dataset are reported. The bold values represent the highest scores for each row.}
\label{tbl:all_results}
\end{table}

\begin{figure}[t]
     \centering
     \begin{subfigure}[b]{0.49\linewidth}
         \centering
         \includegraphics[width=\linewidth]{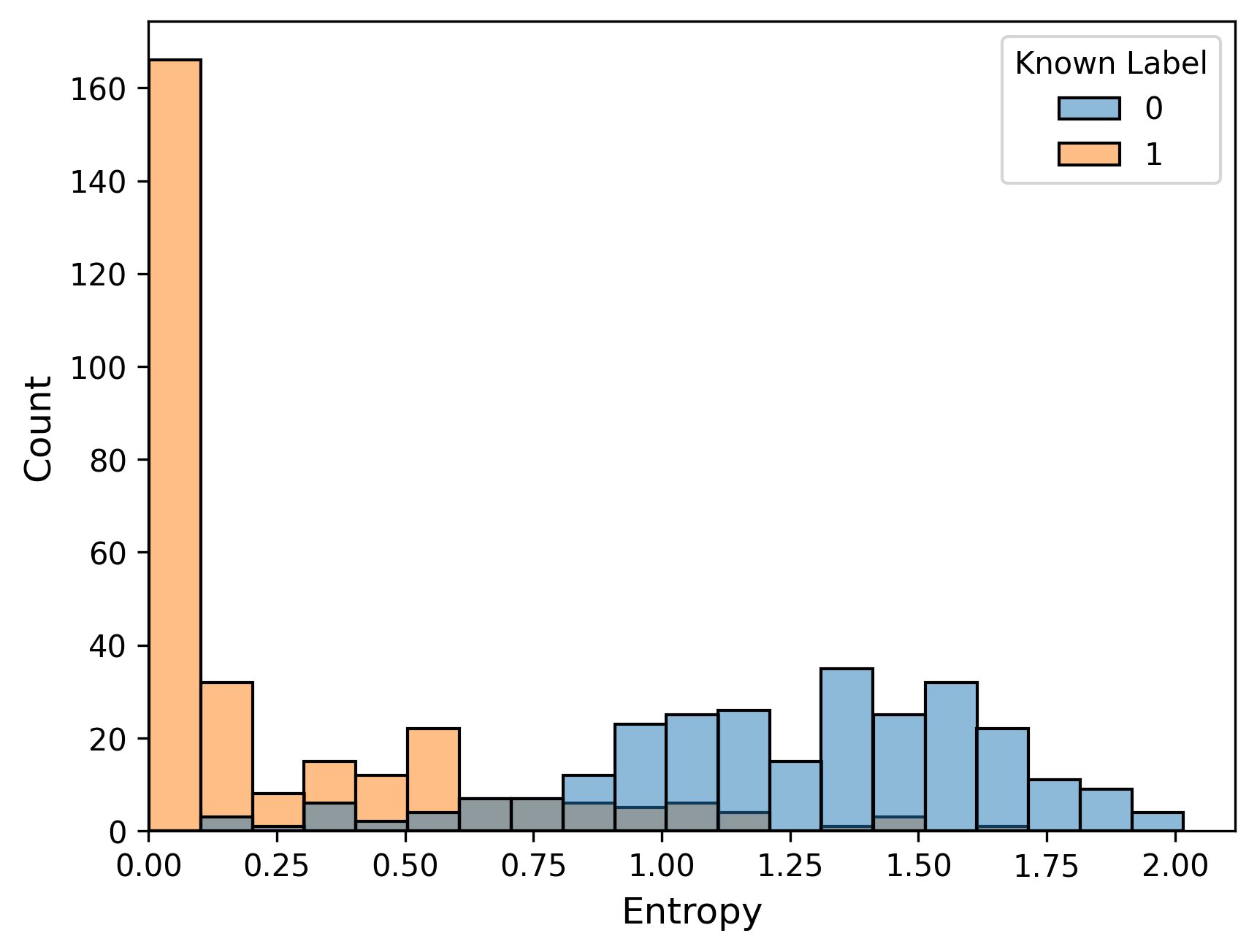}
         \caption{Amazon $\rightarrow$ Webcam in Office dataset.}
     \end{subfigure}
     \hfill
     \begin{subfigure}[b]{0.49\linewidth}
         \centering
         \includegraphics[width=\linewidth]{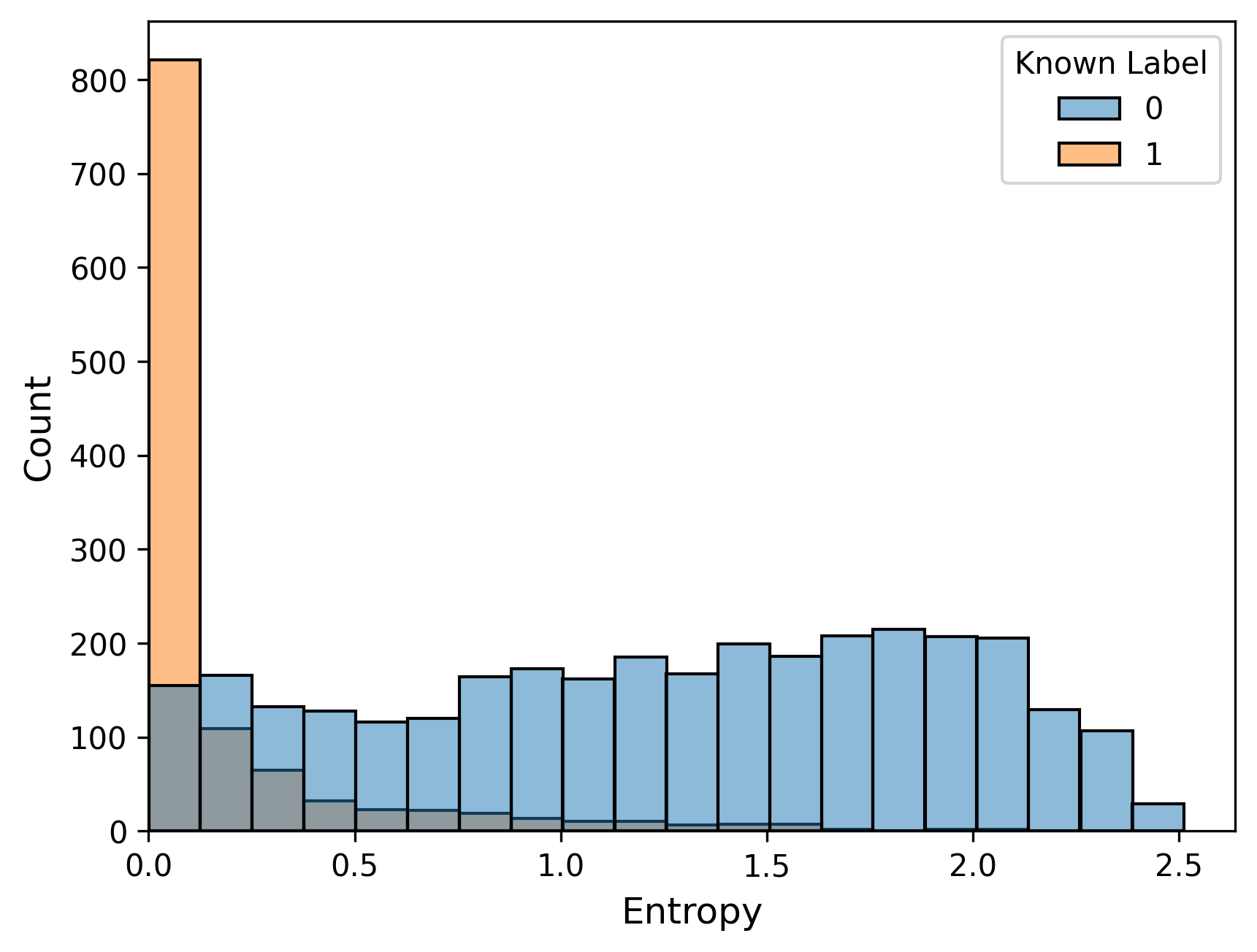}
         \caption{Art $\rightarrow$ Real in Office-Home dataset.}
     \end{subfigure}
    \caption{Histogram of entropy of CLIP's zero-shot prediction on the target known (in orange) and unknown (in blue) samples.}
    \label{fig:hist}
\end{figure}

\begin{figure}[t]
     \centering
     \begin{subfigure}[b]{0.49\linewidth}
         \centering
         \includegraphics[width=\linewidth]{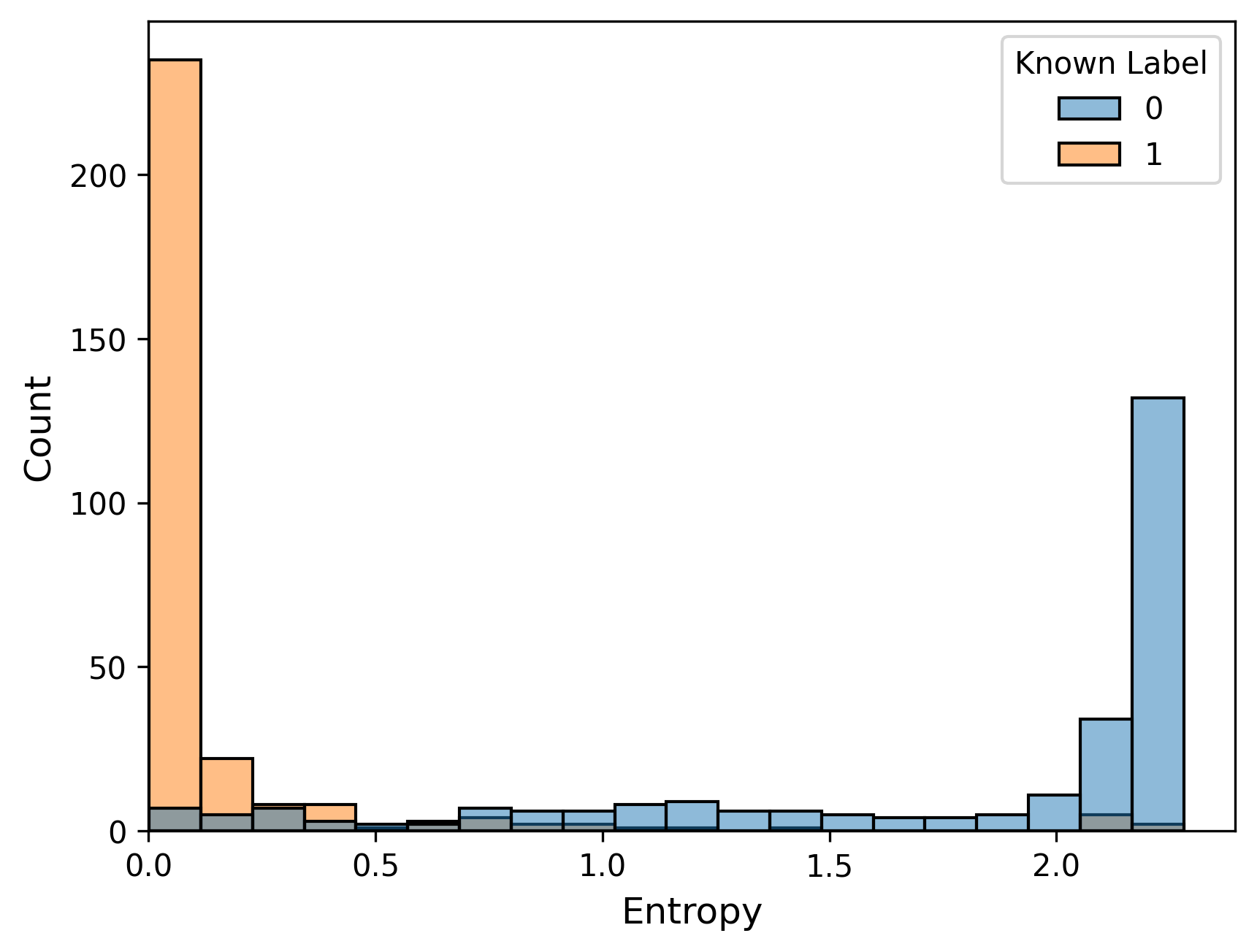}
         \caption{Amazon $\rightarrow$ Webcam in Office dataset.}
     \end{subfigure}
     \hfill
     \begin{subfigure}[b]{0.49\linewidth}
         \centering
         \includegraphics[width=\linewidth]{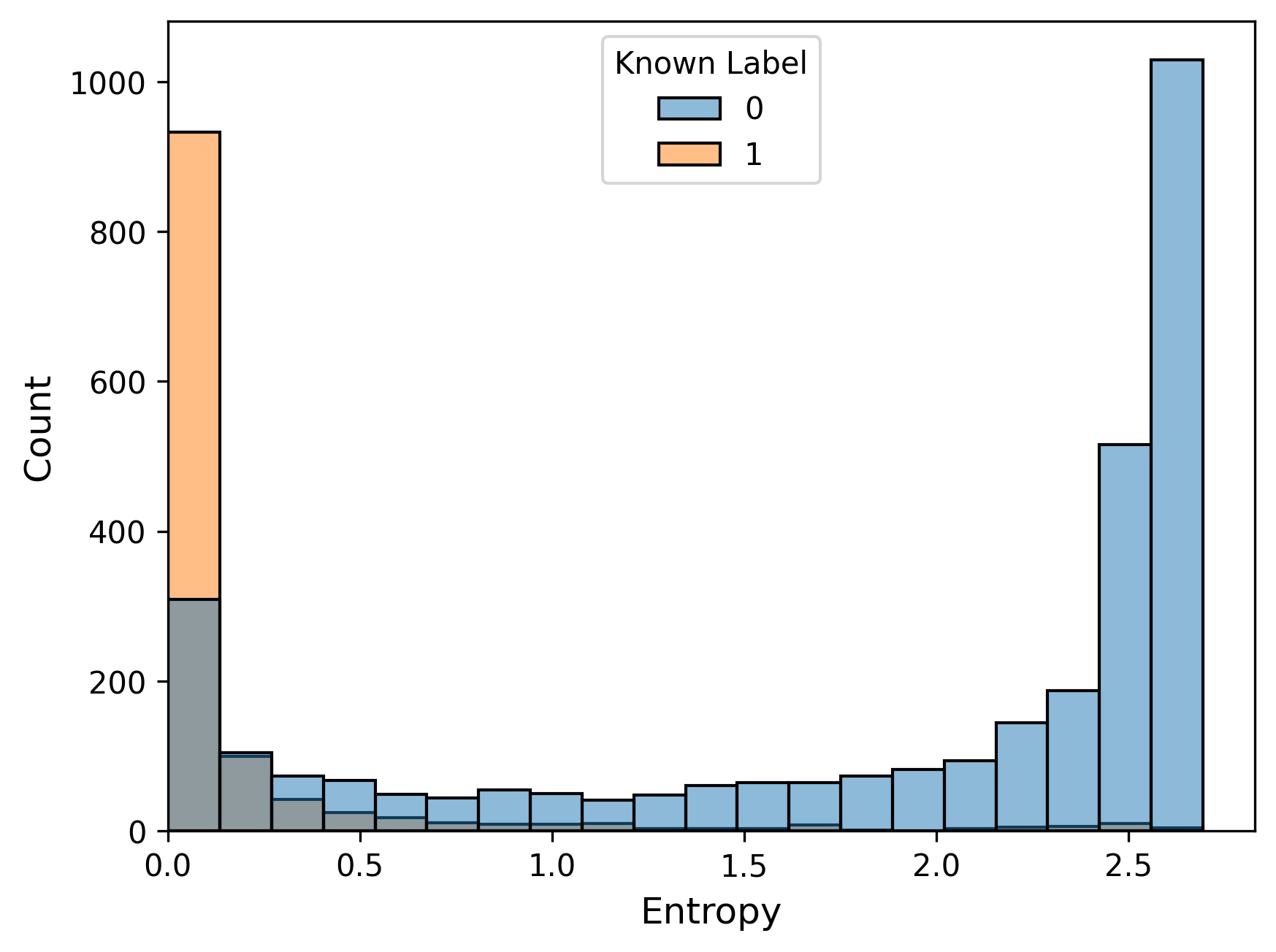}
         \caption{Art $\rightarrow$ Real in Office-Home dataset.}
     \end{subfigure}
    \caption{Histogram of the prediction's entropy obtained from the proposed method on the target known (in orange) and unknown (in blue) samples.}
    \label{fig:hist_2}
\end{figure}

\begin{figure}[t]
 \centering
 \begin{subfigure}[b]{0.49\linewidth}
     \centering
     \includegraphics[width=\linewidth]{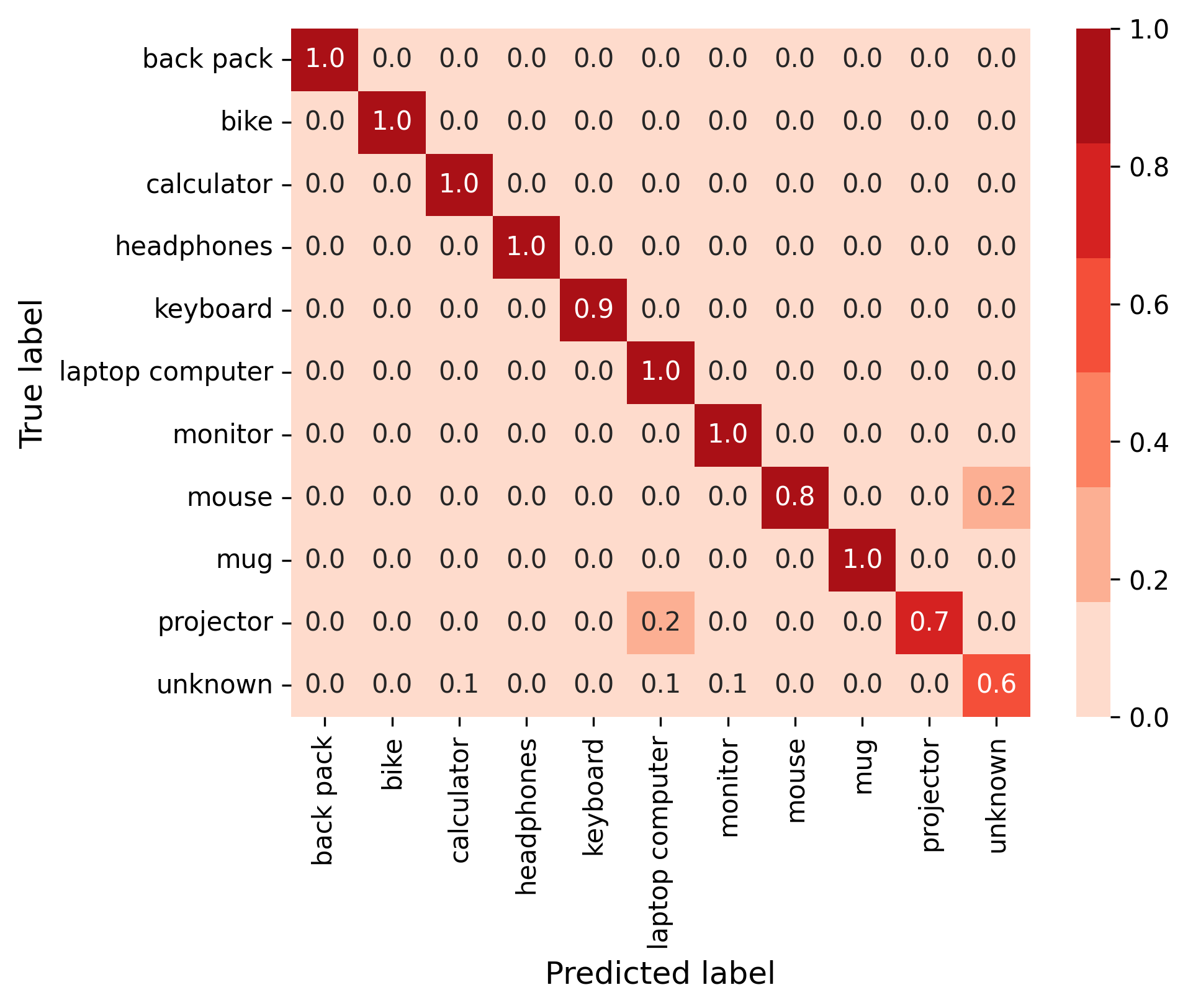}
     \caption{Amazon $\rightarrow$ Webcam in Office dataset.}
 \end{subfigure}
 \hfill
 \begin{subfigure}[b]{0.49\linewidth}
     \centering
     \includegraphics[width=\linewidth]{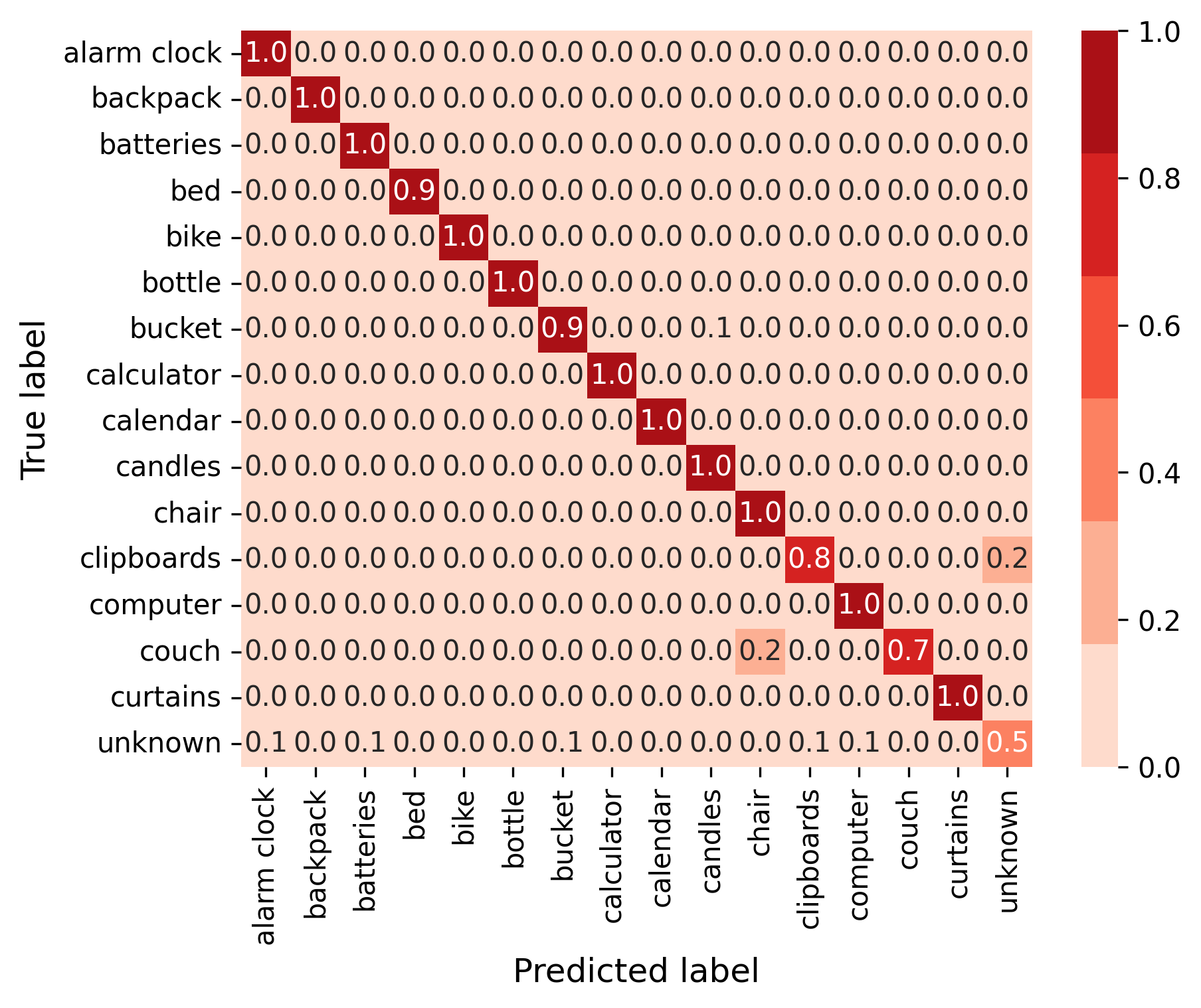}
     \caption{Art $\rightarrow$ Real in Office-Home dataset.}
 \end{subfigure}
 \caption{Confusion matrix of CLIP's zero-shot prediction on the Amazon $\rightarrow$ Webcam dataset.}
 \label{fig:mat}
\end{figure}

\begin{figure}[t]
 \centering
 \begin{subfigure}[b]{0.49\linewidth}
     \centering
     \includegraphics[width=\linewidth]{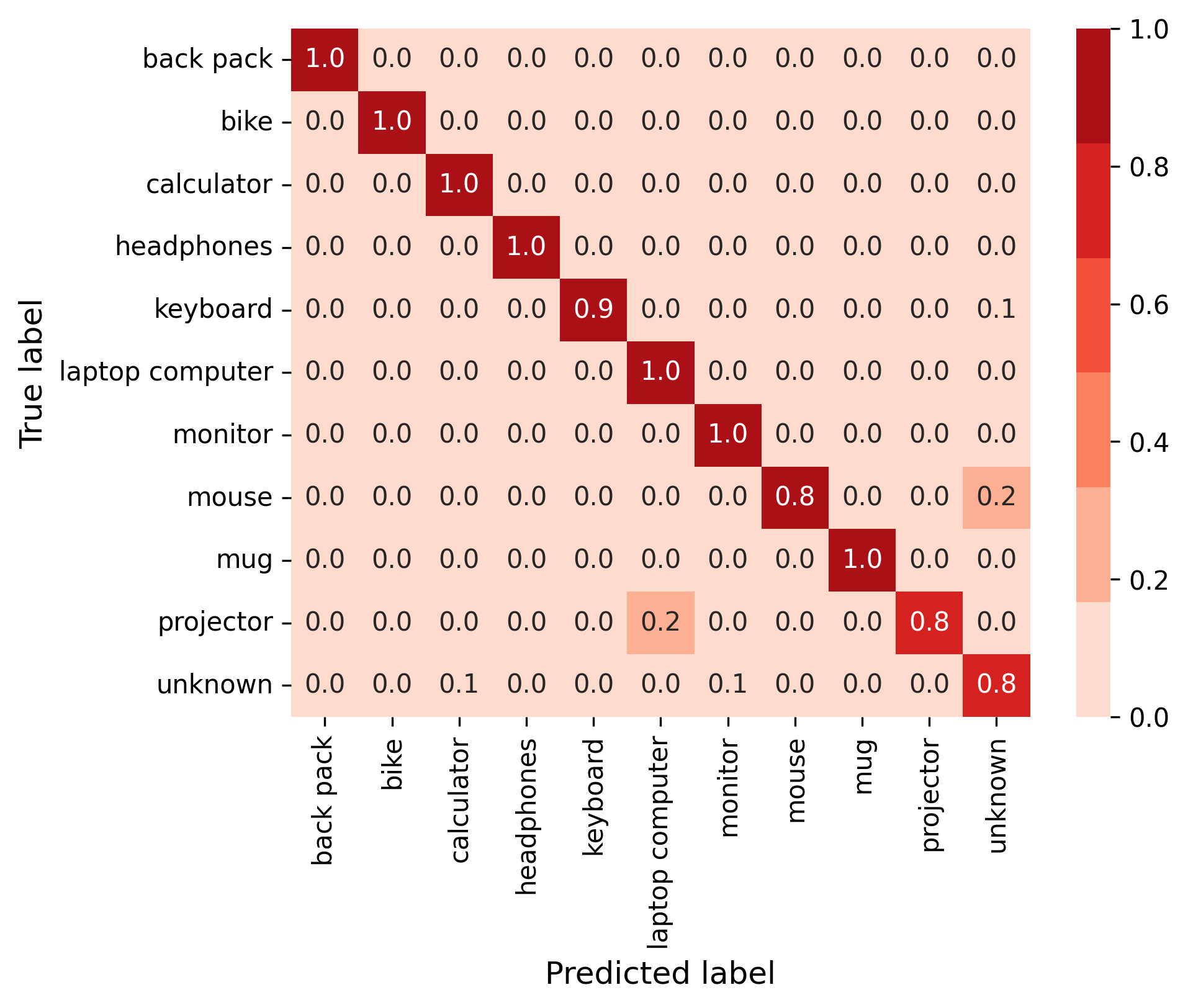}
     \caption{Amazon $\rightarrow$ Webcam in Office dataset.}
 \end{subfigure}
 \hfill
 \begin{subfigure}[b]{0.49\linewidth}
     \centering
     \includegraphics[width=\linewidth]{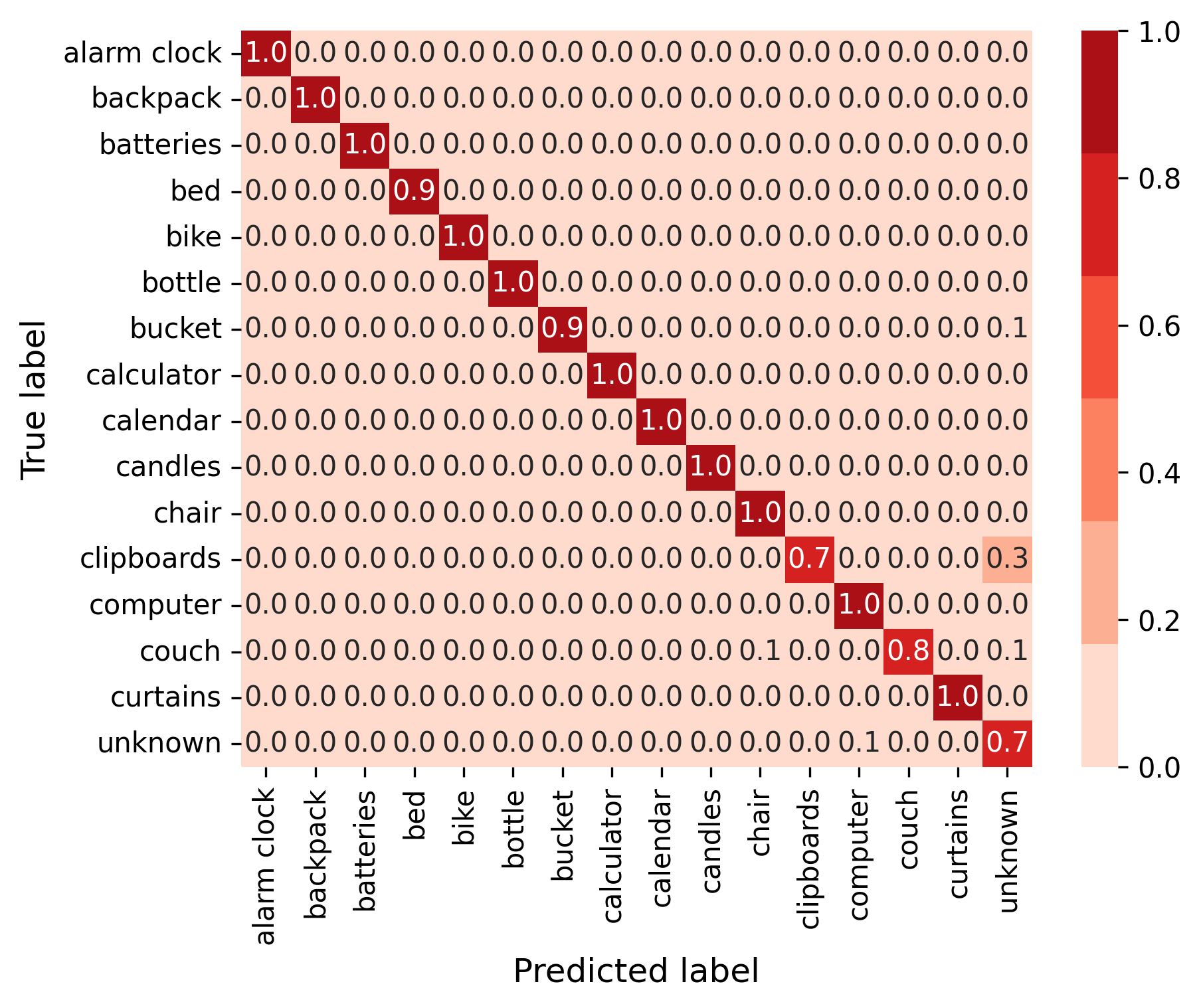}
     \caption{Art $\rightarrow$ Real in Office-Home dataset.}
 \end{subfigure}
 \caption{Confusion matrix of the proposed method on the Amazon $\rightarrow$ Webcam dataset.}
 \label{fig:mat_2}
\end{figure}

\begin{table}[t]
\begin{center}
\tabcolsep = 0.8mm
\begin{tabular}{c|cccccc|c}
\toprule
\multirow{2}{*}{Method}& \multicolumn{6}{c|}{Office}&\multirow{2}{*}{Avg}\\
     & A2D  & A2W  & D2A  & D2W  & W2A  & W2D  &   \\\hline\hline
CLIP    & 81.64 & 75.08 & 72.39 & 75.08 & 72.39 & 81.64 & 76.37 \\ \hline
SO      & 56.59 & 55.81 & 71.04 & 68.06 & 58.58 & 55.37 & 60.91 \\
DANCE   & 83.87 & 77.80 & 77.84 & 84.80 & 68.06 & 74.55 & 77.82 \\
OVA     & 90.29 & 86.30 & 88.28 & 92.66 & 86.14 & 93.75 & 89.57 \\
\rowcolor[gray]{0.90}Ours    & \textbf{93.69} & \textbf{92.05} & 91.59 & \textbf{93.45} & \textbf{90.97} & \textbf{95.00} & \textbf{92.79} \\ \hline
SHOT    & 77.42 & 76.90 & 83.03 & 79.76 & 74.12 & 76.76 & 78.00 \\
OneRing & 89.31 & 85.87 & 88.00 & 92.07 & 90.25 & 93.72 & 89.87 \\
\rowcolor[gray]{0.90}Ours SF & 93.31 & 89.25 & \textbf{91.86} & 92.55 & 90.79 & 93.48 & 91.87 \\
\bottomrule
\end{tabular}
\end{center}
\caption{H-scores (\%) on Office under the ODA and SF-ODA settings.}
\label{tbl:office_results}
\end{table}

\begin{table*}[t]
\begin{center}
\tabcolsep = 1.8mm
\begin{tabular}{c|cccccccccccc|c}
\toprule
\multirow{2}{*}{Method} & \multicolumn{12}{c|}{OfficeHome}                                       & \multirow{2}{*}{Avg}  \\
 & A2C & A2P & A2R & C2A & C2P & C2R & P2A & P2C & P2R & R2A & R2C & R2P &  \\ \hline\hline
CLIP    & 71.17 & 55.15 & 63.94 & 72.57 & 55.15 & 63.94 & 72.57 & 71.17 & 63.94 & 72.57 & 71.17 & 55.15 & 65.71 \\ \hline
SO      & 56.54 & 54.36 & 56.89 & 60.61 & 53.08 & 57.09 & 59.65 & 54.82 & 56.24 & 59.06 & 57.39 & 56.85 & 56.88 \\
DANCE   & 63.10 & 60.28 & 64.78 & 66.62 & 57.98 & 63.62 & 67.55 & 61.92 & 62.81 & 66.87 & 61.86 & 62.59 & 63.33 \\
OVA     & 64.42 & 74.72 & 77.59 & 69.09 & 69.91 & 73.84 & 66.64 & 58.58 & 77.44 & 73.65 & 62.95 & 78.53 & 70.61 \\
\rowcolor[gray]{0.90}Ours    & \textbf{76.66} & 76.02 & 83.41 & 81.51 & 77.28 & 82.44 & 81.93 & 76.48 & 82.65 & \textbf{82.76} & 76.10 & 75.89 & 79.43 \\ \hline
SHOT    & 60.22 & 61.19 & 66.08 & 64.48 & 62.24 & 68.07 & 62.64 & 57.67 & 65.81 & 64.64 & 58.58 & 65.30 & 63.08 \\
OneRing & 60.99 & 69.92 & 73.68 & 67.01 & 67.77 & 71.36 & 67.74 & 57.50 & 74.51 & 67.83 & 61.65 & 71.90 & 67.65 \\
\rowcolor[gray]{0.90}Ours SF & 76.54 & \textbf{79.28} & \textbf{85.46} & \textbf{82.33} & \textbf{78.55} & \textbf{84.83} & \textbf{82.21} & \textbf{76.50} & \textbf{84.59} & 82.09 & \textbf{76.77} & \textbf{78.94} & \textbf{80.67} \\
 \bottomrule
\end{tabular}
\end{center}
\caption{H-scores (\%) on Office-Home under the ODA and SF-ODA settings.}
\label{tbl:home_results}
\end{table*}

\begin{table*}[t]
\begin{center}
\begin{tabular}{c|c|ccccccc|c}
\toprule
\multirow{2}{*}{Method}& \multirow{2}{*}{VisDA} &\multicolumn{8}{c}{DomainNet} \\
& & P2C  & P2R  & C2S  & R2P  & R2C  & R2S  & S2P  &  Avg \\\hline\hline
CLIP    & 79.49 & 66.23 & 67.41 & 64.05 & 67.55 & 66.23 & 64.05 & 67.55 & 66.16 \\ \hline
SO      & 43.57 & 61.51 & 62.41 & 56.32 & 59.99 & 59.90 & 54.42 & 59.40 & 59.14 \\
DANCE   & 67.87 & 60.62 & 58.01 & 55.89 & 59.50 & 59.09 & 53.80 & 59.29 & 58.03 \\
OVA     & 59.80 & 64.27 & 65.18 & 59.58 & 63.79 & 63.67 & 57.28 & 61.81 & 62.23 \\
\rowcolor[gray]{0.90}Ours    & 80.68 & 78.26 & 84.48 & \textbf{72.26} & \textbf{74.45} & 78.18 & \textbf{72.07} & \textbf{73.89} & \textbf{76.23} \\ \hline
SHOT    & 47.08 & 61.55 & 62.82 & 55.45 & 58.36 & 61.54 & 55.40 & 55.36 & 58.64 \\
OneRing & 51.21 & 61.40 & 66.17 & 53.85 & 59.38 & 59.53 & 52.19 & 56.72 & 58.46 \\
\rowcolor[gray]{0.90}Ours SF & \textbf{83.81} & \textbf{78.45} & \textbf{84.96} & 72.06 & 73.58 & \textbf{78.89} & 71.95 & 73.03 & 76.13 \\
\bottomrule
\end{tabular}
\end{center}
\caption{H-scores (\%) on VisDA and DomainNet under the ODA and SF-ODA settings.}
\label{tbl:domain_results}
\end{table*}

\textbf{Main results.}
\Tref{tbl:all_results} shows the ODA and SF-ODA results on each dataset. We compared our method (non-source-free and source-free) with ODA methods, DANCE and OVA, and SF-ODA methods, SHOT and OneRing. It is noticeable that the proposed method outperformed other existing methods by a large margin. It is also surprising to find that the zero-shot prediction of CLIP is comparable to some ODA and SF-ODA methods in Office and Office-Home datasets. In VisDA and DomainNet, we found that CLIP performs better than all the compared methods without any fine-tuning. We considered that this is because the classes in these two datasets are more coarse-grained and the domains are more common in the training data of CLIP, \eg, Painting, Clipart, Sketch, and Real. These results demonstrate that with the power of large-scale pre-training containing training data from multiple domains, existing vision-language foundation models like CLIP are enough to cover the existing datasets for ODA in image classification.

\textbf{Comparison between zero-shot CLIP and the proposed method.}
To further investigate the performance of CLIP, In~\Fref{fig:hist}, we show the histograms of the entropy of CLIP's zero-shot predictions on the target known and unknown samples. We can find that there are some overlaps between the known and unknown samples. Especially, in the Art $\rightarrow$ Real task of the Office-Home dataset, a lot of unknown samples have small entropy, which leads to the low performance of the detection of unknown samples. We plot the histograms of the proposed method in~\Fref{fig:hist_2} and the proposed method can separate the known and unknown samples better than CLIP by the threshold $\delta$. We calculate the area under the receiver operating characteristic curve (AUROC) by regarding the detection of unknown samples as a binary classification. The AUROC of CLIP and the proposed method in Amazon $\rightarrow$ Webcam are 95.91\% and 98.18\%, respectively. For Art $\rightarrow$ Real, the AUROC of CLIP and the proposed method is 93.82\% and 94.71\%.

Furthermore, we plot the confusion matrix of CLIP in~\Fref{fig:mat} in the task Amazon $\rightarrow$ Webcam of Office dataset and Art $\rightarrow$ Real task of Office-home dataset. We can see that although the classification of known samples is almost correct, there are some unknown samples predicted as other known classes. In~\Fref{fig:mat_2}, we also plot the confusion matrix of the proposed method. We can find that the proposed method achieves better performance in the detection of unknown samples with the help of the ODA model trained on the source domain.

\textbf{Detail results on each dataset.}
The results of each task in the Office dataset are shown in~\Tref{tbl:office_results}, which compares the classification results of ODA and SF-ODA obtained via the proposed method with existing state-of-the-art ODA and SF-ODA methods. In terms of ODA, although OVA demonstrated superior performance compared to other existing techniques, our approaches surpassed OVA in most tasks. Furthermore, in the SF-ODA setting, although there is no access to the source data, OneRing achieved good results, which is a little better than that of OVA. Meanwhile, the source-free version of the proposed approach outperformed existing methods by a considerable margin and achieved similar performance to the ODA version. 

The results of each task in Office-Home, VisDA and DomainNet are shown in~\Tref{tbl:home_results} and~\Tref{tbl:domain_results}. The performance ranking of each existing method and the proposed method is similar to the Office dataset, where OVA performs best in most datasets but the proposed method outperforms it by a large margin. As mentioned in the main results, CLIP shows strong results on VisDA and DomainNet.

\subsection{Ablation Study}
Variants of the proposed method were evaluated using the Office dataset, for further exploration of the efficacy of the proposed method. The following variants were studied. (1) ``Ours w/o $\mathcal{L}_{s}$'' is a variant that does not train with source data in \Eref{eq:L_s}. (2) ``Ours w/o $\mathcal{L}_{ent}$'' is a variant that does not use entropy separation on target samples in \Eref{eq:L_ent}. (3) ``Ours w/ $\mathcal{L}_{kwn}$'' is a variant that does not use the prediction of CLIP as the pseudo label in \Eref{eq:L_kwn}. (4) ``Ours w/o $\mathcal{L}_{unk}$'' is the variant that does not maximize the entropy of unknown samples detected by CLIP in \Eref{eq:L_unk}. \Tref{tbl:ab} reveals that the version of our approach that uses all the losses outperforms other variants in all settings. Specifically, the most important component for our method is $\mathcal{L}_{unk}$, and $\mathcal{L}_{kwn}$ is also necessary to achieve higher performance, which shows the importance of the CLIP's guidance.
 
\begin{table}[t]
\centering
\begin{center}
\begin{tabular}{c|cccc}
\toprule
Method & OF & OH & VD & DN \\ \hline \hline 	
\rowcolor[gray]{0.90}Ours & \textbf{92.79} & \textbf{79.43} & \textbf{80.68} & \textbf{76.23} \\ \hline 
w/o $\mathcal{L}_{s}$&90.04 & 71.86 & 78.10 & 74.79 \\
w/o $\mathcal{L}_{ent}$&90.05 & 78.74 & 77.32 & 72.35 \\
w/o $\mathcal{L}_{kwn}$&87.87 & 72.98 & 72.12 & 70.55 \\
w/o $\mathcal{L}_{unk}$&82.67 & 66.98 & 71.63 & 69.14\\
\bottomrule
\end{tabular}
\end{center}
\caption{H-score (\%) of ablation study tasks on each dataset.}
\label{tbl:ab}
\end{table}

\section{Limitations and Future work}
We proposed a simple method to use CLIP for enhancing the performance of the ODA model, but we believe there could be more efficient methods to adopt CLIP in ODA. Fine-tuning CLIP directly can also be considered, but the computationally cost needs to be concerned, and preventing the fine-tuning process from causing the model to overfit to the source
domain is necessary. It is certainly intriguing to develop a more complicated method for ODA with CLIP, which is our future work.

\section{Conclusion}
In this paper, we proposed a method to enhance the performance of ODA using CLIP. To address the transfer of knowledge across domains with different class sets, including unknown classes, the proposed approach uses the zero-shot predictions of CLIP to enhance the performance of ODA models. We calculate the entropy of the outputs of the ODA model and the predictions of CLIP on the target domain to identify known and unknown samples. The method then distills the predictions of CLIP to the ODA model to help the adaptation of target known samples and separates unknown samples by maximizing the entropy of the ODA model. We demonstrated the effectiveness of the proposed method on various ODA benchmarks and showed that it outperformed current ODA and SF-ODA methods.

\section*{Acknowledgements} 
This work was supported by JST AIP Acceleration Research Grant Number JPMJCR22U4 and JSPS KAKENHI Grant Number 20J22372.

{\small
\bibliographystyle{ieee_fullname}
\bibliography{egbib}
}

\end{document}